\documentclass[sigconf]{acmart}

\usepackage{booktabs}
\usepackage{graphicx}
\usepackage{amsmath}
\usepackage{multirow}
\usepackage{xcolor}
\usepackage{microtype}

\usepackage[subtle]{savetrees}
\acmConference[BALANCES'26]{6th ACM International Workshop on Big Data and Machine Learning for Smart Buildings and Cities}{June 22, 2026}{Banff, Alberta, Canada}
\acmYear{2026}
\copyrightyear{2026}

\begin{document}

\title{Uncertainty-Aware Transfer Learning for Cross-Building Energy
Forecasting: Toward Robust and Scalable District-Level Energy Management}

\author{Shadmehr Zaregarizi}
\affiliation{%
  \institution{Politecnico di Torino}
  \city{Turin}
  \country{Italy}}
\email{shadmehr.zaregarizi@studenti.polito.it}

\author{Khashayar Yavari}
\affiliation{%
  \institution{Politecnico di Torino}
  \city{Turin}
  \country{Italy}}
\email{khashayar.yavari@studenti.polito.it}

\begin{abstract}
Scaling data-driven energy forecasting to district level requires models
that can be re-used across buildings with minimal target-domain data and
honest uncertainty estimates. We present an uncertainty-aware transfer
learning (TL) framework for cross-building energy forecasting based on the
Temporal Fusion Transformer (TFT), evaluated on a newly released
high-resolution real sub-meter dataset: an educational building at
Aalborg University, Denmark (source) and the multi-typology NEST building
at EMPA, Switzerland (target). We introduce the \emph{Transfer Robustness
Index} (TRI), an architecture-agnostic metric for quantifying
generalization quality across domain gaps. A four-strategy layer-freezing
ablation shows that \emph{Probe-Only} fine-tuning, updating only 455
output-layer parameters out of 806K, achieves the best transfer quality
(TRI = 3,097), outperforming full fine-tuning and suggesting that TFT
encoders learn transferable temporal representations. Monte Carlo Dropout
yields a prediction interval coverage probability of 93.2\%, close to the
nominal 95\% target. A data-scarcity analysis further shows monotonic
improvement with increasing target-domain data, providing practical
guidance for district energy deployment.
\end{abstract}


\maketitle

\section{Introduction}

Sub-meter sensing, IoT, and big-data analytics now make data-driven
energy management feasible at district scale. Yet re-using a model
trained on one building for another in the same district remains an
open challenge: \emph{domain gap} (typology, climate, occupancy,
system configuration), \emph{uncertainty blindness} (no confidence
bounds on an unseen building), and \emph{data hunger} (new buildings
rarely have enough history to train from scratch) compound at once.
Prior building-energy forecasting work has largely been single-building
or simulation-based; transfer learning (TL) has been explored for HVAC
control and occupancy. While recent advances in 2024 and 2025 have
successfully leveraged Transformer-based attention mechanisms for
complex sequence modeling in energy systems~\cite{timesfm2024, moirai2024},
their systematic application to \emph{electricity demand forecasting
with multi-energy-vector inputs across distinct real buildings},
combined with uncertainty quantification and data-scarcity guidelines,
is underreported. Furthermore, no standardized metric exists in the
literature to compare TL generalization quality across diverse studies.

This paper addresses these gaps with four contributions: (i) a
\textbf{TFT-based cross-building TL framework} using multi-energy-vector
covariates (electricity, heating, DHW), evaluated as a cross-country
case study on the AAU (Denmark) and NEST (Switzerland) buildings;
(ii) the \textbf{Transfer Robustness Index (TRI)}, an
architecture-agnostic metric for standardized evaluation of TL
generalization in building energy forecasting; (iii) a
\textbf{four-strategy layer-freezing ablation} revealing that
Probe-Only fine-tuning (455 parameters) yields superior results
($TRI = 3{,}097$), suggesting that TFT encoders learn domain-agnostic
temporal abstractions; and (iv) \textbf{calibrated uncertainty
quantification} via MC Dropout ($PICP = 93.2\%$) alongside data-scarcity
analysis for practical district-level deployment.

\section{Related Work}

\textbf{Building energy forecasting.} Deep learning methods, including
LSTMs, Transformers, and hybrid architectures, perform well on
single-building forecasting. The Temporal Fusion Transformer
(TFT)~\cite{tft2021} is notable for its gated residual networks and
multi-head attention, providing interpretable multi-horizon forecasts.
Recent 2024--2026 advances have explored time-series foundation
models~\cite{timesfm2024} and large pre-trained sequence
models~\cite{moirai2024} for zero-shot and few-shot transfer.
Despite this, most applied building-energy studies remain focused on
per-building training or rely on simulation-based evaluation.

\textbf{Transfer learning for buildings.} TL addresses data scarcity via
domain adaptation or fine-tuning. In building energy applications,
recent work on transfer learning with Transformer architectures
\cite{spencer2025transfer} has evaluated data-centric transfer and
fine-tuning strategies for building energy consumption forecasting.
However, systematic evaluation of TL forecasting across
\textbf{structurally diverse} real buildings integrated with uncertainty
quantification (UQ) remains underreported, and no standardized
robustness metric is in common use.

\textbf{Uncertainty quantification.} Monte Carlo Dropout~\cite{gal2016dropout}
provides practical Bayesian inference, while conformal prediction
offers distribution-free coverage guarantees. UQ remains underexplored
in TL contexts, where epistemic uncertainty---the uncertainty stemming
from the model's lack of knowledge---naturally grows when the model is
deployed to an unseen target building.

\section{Dataset}

We use the high-resolution building dataset published on Zenodo in
March 2026~\cite{dataset2026}, a 4.6\,GB open-access resource
(CC~BY~4.0) comprising two pilot buildings.

\textbf{Source domain --- AAU Pilot (Denmark).} An educational building
at Aalborg University instrumented with sub-meter electricity (69
channels), space heating (20 channels), and domestic hot water (DHW,
9 channels), May 2022 -- July 2024 ($\approx$2~years) at
1--10~minute resolution.

\textbf{Target domain --- NEST Pilot (Switzerland).} The EMPA NEST
building hosts four research units that span very different
typologies: \emph{DFAB} (a 3D-printed concrete residential unit),
\emph{HiLo} (a lightweight, ultra-low-energy roof and facade
research module), \emph{Sprint} (a modular wood office unit), and
\emph{UMAR} (a residential unit built from reused materials).
Data spans October 2023 -- July 2024 ($\approx$9~months). The two
sites differ in climate (Danish maritime vs.\ Swiss alpine), scale,
and use, maximizing domain heterogeneity.

\textbf{Signals used.} Aggregate building electricity is the
forecasting target; sub-meter electricity, space-heating, and DHW
channels enter as time-varying covariates, providing the model with
multi-energy-vector context.

\paragraph{Preprocessing} Raw data undergoes: (1) resampling to a
hourly grid; (2) short-gap filling by linear interpolation for
outages $\leq 4$~h, while longer gaps are excluded from the loss;
(3) conversion of cumulative meter readings to hourly consumption
by finite differences; (4) removal of channels with $>30\%$ missing
values; and (5) independent min--max normalization per building, fit
only on training data to prevent leakage. Final tensor shapes are
AAU $13{,}244 \times 254$ (train) and NEST $4{,}965 \times 312$
(fine-tune) / $1{,}437 \times 312$ (test).

\section{Methodology}

\subsection{Base Model and Inputs}

\textbf{TFT background.} The Temporal Fusion Transformer
(TFT)~\cite{tft2021} is an encoder--decoder architecture for
multi-horizon multivariate forecasting. Its main blocks are a
\emph{Variable Selection Network} for per-step input importance,
\emph{Gated Residual Networks} for skip connections with learnable
gating, an \emph{LSTM encoder--decoder} on the lookback and forecast
windows, and \emph{multi-head self-attention} on top of the LSTM
states for long-range dependencies. A quantile loss yields a
conditional predictive distribution rather than a single-point
prediction.

\textbf{Inputs.} (i) \emph{Static covariates}: building identifier;
(ii) \emph{time-varying known}: cyclical time features (hour-of-day,
day-of-week, month) plus local weather \emph{observations} from the
dataset, treated as known because we feed observed weather rather than
operationally forecasted weather; and (iii) \emph{time-varying unknown}:
historical sub-meter consumption, including electricity, heating, and
DHW.

\textbf{Forecast target.} A single scalar series: the aggregate
hourly building-level electricity demand, computed as the sum of
electricity sub-meters and normalized, over the next 24~h. Heating
and DHW are not predicted; they enter the encoder as auxiliary
covariates that provide multi-energy-vector context.

\textbf{Configuration.} Hidden size 64, 4 attention heads, dropout
0.1, encoder lookback 168~h, forecast horizon 24~h, seven-quantile
loss ($q\in\{0.02,0.1,0.25,0.5,0.75,0.9,0.98\}$); 806K parameters.
Adam ($lr=10^{-3}$), gradient clipping ($\|\nabla\|\leq0.1$), early
stopping with patience 5.

\subsection{Transfer Learning Pipeline}

\emph{Layer-freezing} adaptation re-uses a model trained on a source
domain by holding the parameters of selected layers fixed during
target-domain fine-tuning, while updating only the remaining layers.
This biases the optimizer toward preserving features learned from
the abundant source data and reduces the number of effective
parameters that have to be re-learned from limited target data.

The pre-trained AAU model serves as the source for four
layer-freezing strategies, each fine-tuned on the NEST fine-tuning
split for up to 20 epochs at $\text{lr}=10^{-4}$. Sub-meter
channels present in NEST but not in AAU are zero-padded to keep
the input dimensionality fixed.

\begin{itemize}\itemsep1pt
  \item \textbf{Full Fine-Tuning (FF):} all 806K parameters updated.
        Maximum adaptation capacity but high risk of catastrophic
        forgetting on limited target data.
  \item \textbf{Partial Fine-Tuning (PF):} input embedding layers
        frozen; encoder, decoder, and output head updated.
  \item \textbf{Probe Only (PO):} all layers frozen except the final
        output projection (455 parameters; see breakdown below).
        Tests whether source representations are directly re-usable
        through a fresh output head.
  \item \textbf{Progressive Unfreezing (PU):} encoder and embeddings
        frozen (567K); decoder and output head updated (239K).
        Preserves low-level temporal representations while adapting
        higher-level sequence modeling.
\end{itemize}

\textbf{How 455 ``probe'' parameters arise.} The TFT output head is
a linear projection from the decoder hidden state ($d{=}64$) to one
quantile per horizon step. With 7 quantiles this gives a
$64 \times 7$ weight matrix (448 parameters) plus a 7-dim bias,
for a total of $448 + 7 = 455$ trainable parameters in PO.

\textbf{Baselines:} (B1) Seasonal \textbf{Persistence}: next 24\,h
equals the same period 24\,h prior; (B2) \textbf{LSTM trained from scratch}
on AAU and evaluated on NEST (327K parameters, 30 epochs); and (B3)
\textbf{Direct Transfer}: TFT applied to the target domain without
fine-tuning.

\subsection{Transfer Robustness Index}

We define the Transfer Robustness Index as:
\begin{equation}
  \mathrm{TRI} = \frac{\mathrm{MAE}_{\mathrm{source\,val}}}
                      {\mathrm{MAE}_{\mathrm{target\,test}} + \epsilon},
  \quad \epsilon = 10^{-8}
  \label{eq:tri}
\end{equation}
where $\mathit{MAE}_{\mathrm{src,val}}=15.745$ on the AAU validation
set and $\mathit{MAE}_{\mathrm{tgt,test}}$ is the error on the NEST
test set after transfer. Both terms are computed on independently
min--max normalized scales using per-building training statistics, so
$\mathit{TRI}$ quantifies relative generalization improvement within
each domain's own scale rather than an absolute cross-building error
comparison. $\mathit{TRI}>1$ means the model generalizes better on
the target than on the source validation set; large values, such as
$\approx 3{,}000$, stem from the absolute-error disparity between the
two normalized series. A decomposed variant
$\mathit{TRI}=\mathit{TRI}_{\mathrm{climate}}\times\mathit{TRI}_{\mathrm{typology}}$
is possible when intermediate sites become available.

\subsection{Uncertainty Quantification}

Monte Carlo (MC) Dropout~\cite{gal2016dropout} is applied by enabling
all dropout layers during inference and running $N=50$ stochastic
forward passes on the best fine-tuned model. The point estimate is
the mean across passes; 95\% prediction intervals use the 2.5th and
97.5th percentiles. Calibration is measured by PICP, the fraction of
true values within the interval, and mean interval width (MIW).

\textbf{Heterogeneity vs.\ uncertainty.} The two goals are
addressed by separable mechanisms. \emph{Heterogeneity}, meaning
different buildings, climates, and uses, is handled at training time
by the source--target TL pipeline, the building-specific input
embedding with zero-padding alignment, and the layer-freezing
strategy that preserves transferable encoder representations.
\emph{Uncertainty}, meaning lack of confidence on a new building, is
handled at inference by the quantile-loss head plus MC Dropout,
yielding calibrated prediction intervals. TRI then summarizes how
robustly the heterogeneity machinery generalizes, while PICP and MIW
summarize the calibration of the uncertainty machinery.

\textbf{Data-scarcity protocol.} The base TFT is fine-tuned, using
full fine-tuning, on four progressively larger NEST subsets taken
chronologically from the fine-tuning pool: 2~weeks (336~h),
1~month (720~h), 3~months (2,160~h), and all available data
(4,965~h). Each subset is evaluated on the same fixed NEST test set.

\section{Results}

\subsection{Transfer Learning Performance}

\begin{figure}[t]
  \centering
  \includegraphics[width=0.78\columnwidth]{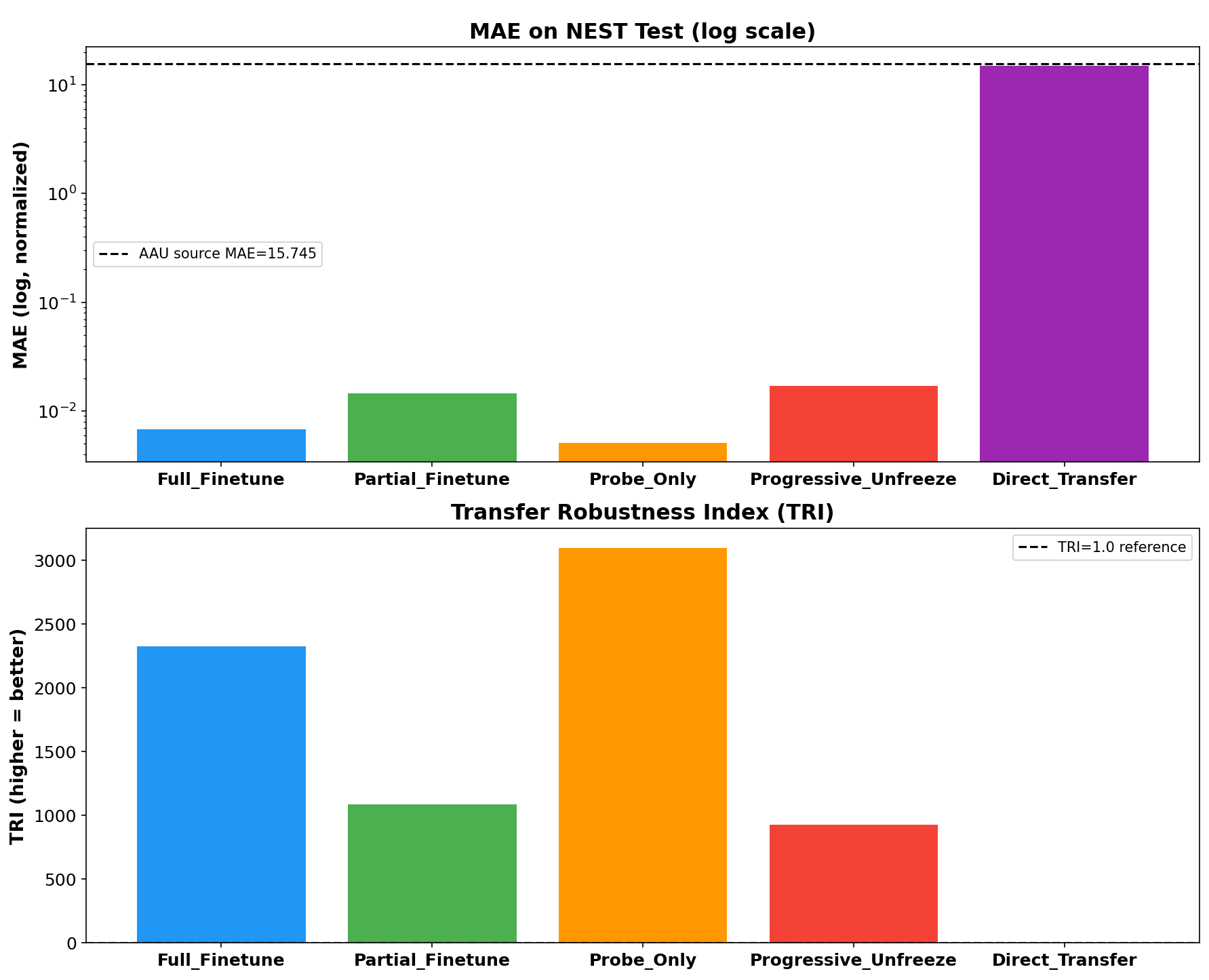}
  \Description{Two-panel bar and line chart. Left panel shows MAE for different TL strategies; right panel shows TRI. Probe Only shows the lowest MAE among TL strategies and the highest TRI.}
  \caption{Transfer learning results on NEST test set. All fine-tuned strategies vastly outperform Direct Transfer.}
  \label{fig:transfer}
\end{figure}

Table~\ref{tab:results} reports all model results on the NEST
electricity test set. Figure~\ref{fig:transfer} visualizes MAE
and TRI across strategies.

\begin{table}[t]
\caption{Forecasting results on NEST electricity test set. \textbf{Bold} indicates best fine-tuned model. TRI computed via Eq.~\ref{eq:tri}; $\mathit{MAE}_{\mathrm{src}}=15.745$.}
\label{tab:results}
\footnotesize
\setlength{\tabcolsep}{4pt}
\begin{tabular}{lrrrr}
\toprule
\textbf{Model} & \textbf{MAE} & \textbf{RMSE} & \textbf{$R^2$} & \textbf{TRI} \\
\midrule
Persistence          & 0.0044 & 0.0352 & $-0.862$ & --     \\
LSTM (scratch)       & 0.0060 & 0.0266 & $-0.005$ & --     \\
Direct Transfer      & 15.037 & 15.038 & $-364{,}644$ & 1.05 \\
\midrule
FF (Full Finetune)   & 0.0068 & 0.0256 & $-0.056$ & $2{,}323$  \\
PF (Partial FT)      & 0.0145 & 0.0275 & $-0.218$ & $1{,}085$  \\
\textbf{PO (Probe Only)} & \textbf{0.0051} & \textbf{0.0255} & \textbf{$-0.046$} & \textbf{$3{,}097$} \\
PU (Prog. Unfreeze)  & 0.0170 & 0.0302 & $-0.468$ & 928    \\
\bottomrule
\end{tabular}
\end{table}

Direct Transfer collapses completely (MAE\,=\,15.037,
TRI\,=\,1.05), confirming that zero-shot cross-building transfer
is infeasible at this domain gap. All fine-tuned variants achieve
MAE reductions exceeding 99.9\% relative to Direct Transfer.
\textbf{Among TL strategies, Probe Only performs best}: MAE\,=\,0.0051,
RMSE\,=\,0.0255, $R^2$\,=\,$-$0.046, TRI\,=\,3,097.
Updating only 455 output-layer parameters outperforms updating all
806K parameters (Full Finetune, TRI\,=\,2,323), suggesting that the
TFT encoder learns domain-agnostic temporal abstractions, including
diurnal and weekly energy consumption patterns, that transfer directly
across building typologies and climates, while only the final output
mapping requires domain-specific adaptation. Negative $R^2$ values
reflect the difficulty of capturing high-frequency stochastic spikes;
therefore, $\mathit{MAE}$ and $\mathit{TRI}$ serve as the primary
relative indicators of improvement and robustness.

\subsection{Uncertainty Quantification}

\begin{figure}[t]
  \centering
  \includegraphics[width=0.78\columnwidth]{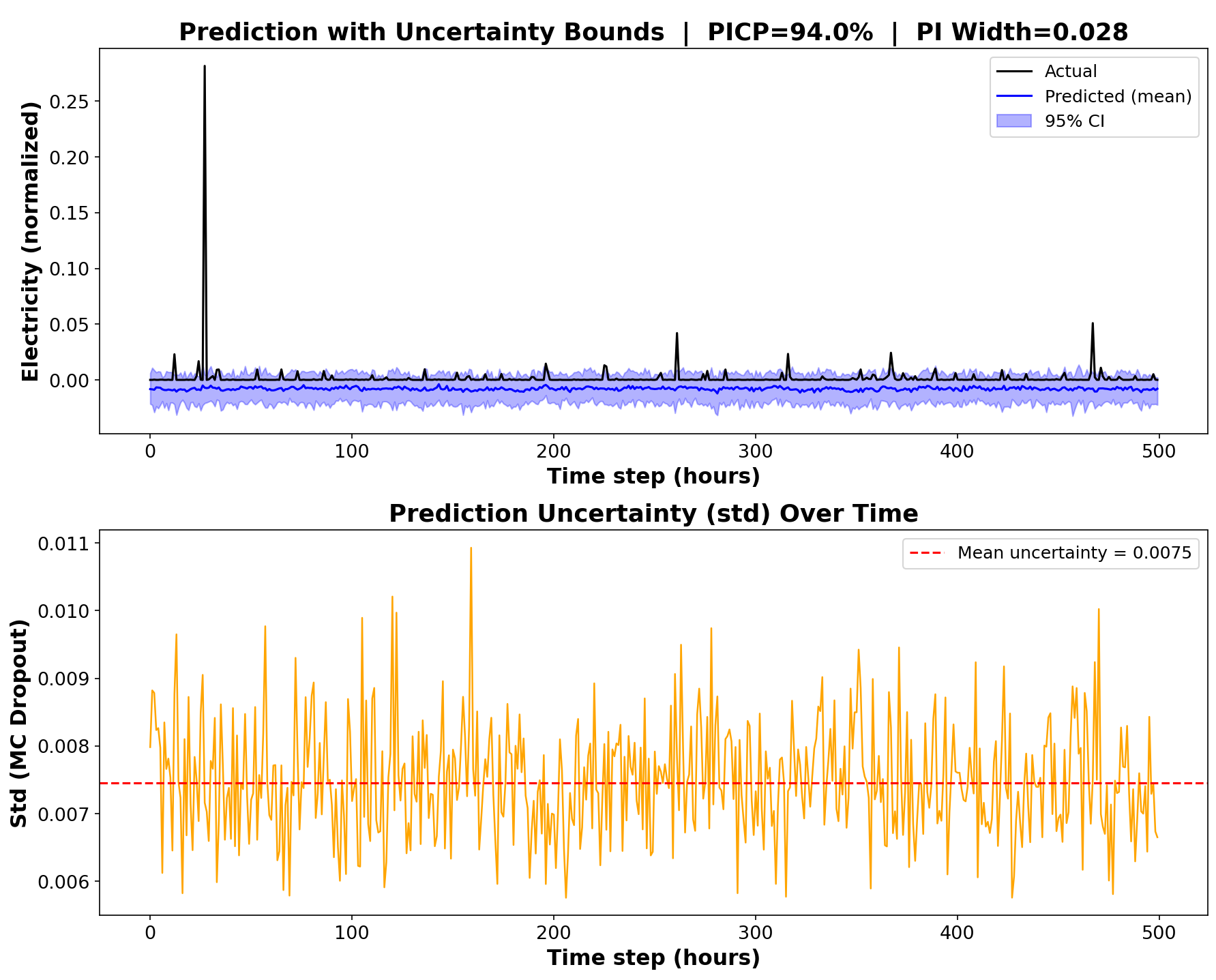}
  \caption{MC Dropout ($N=50$) prediction intervals on the NEST test set using the Probe-Only model. Top: ground truth, mean prediction, and 95\% interval in normalized units. Bottom: per-step prediction standard deviation. The interval widens near the hour-50 surge, indicating increased uncertainty in unfamiliar high-variability regimes. PICP\,=\,93.2\%.}
  \label{fig:uncertainty}
\end{figure}

MC Dropout on the Probe-Only model yields PICP\,=\,\textbf{93.2\%}
(nominal: 95\%), with mean interval width MIW\,=\,0.028 and
MAE\,=\,0.0097 for the mean prediction. Figure~\ref{fig:uncertainty}
shows prediction intervals over a 500-hour window of the NEST
test period; the values shown are min--max normalized in the
$[0,1]$ range used for training.

The near-nominal PICP shows the intervals are well calibrated,
with roughly 19 of every 20 hourly observations falling inside the
band. The intervals widen locally around the hour-50 surge, indicating
that MC Dropout reports \emph{higher} uncertainty in unfamiliar regimes
rather than producing a uniform band. The tight average MIW confirms
that the intervals are informative rather than trivially wide.

\subsection{Data Scarcity Analysis}

\begin{figure}[t]
  \centering
  \includegraphics[width=0.78\columnwidth]{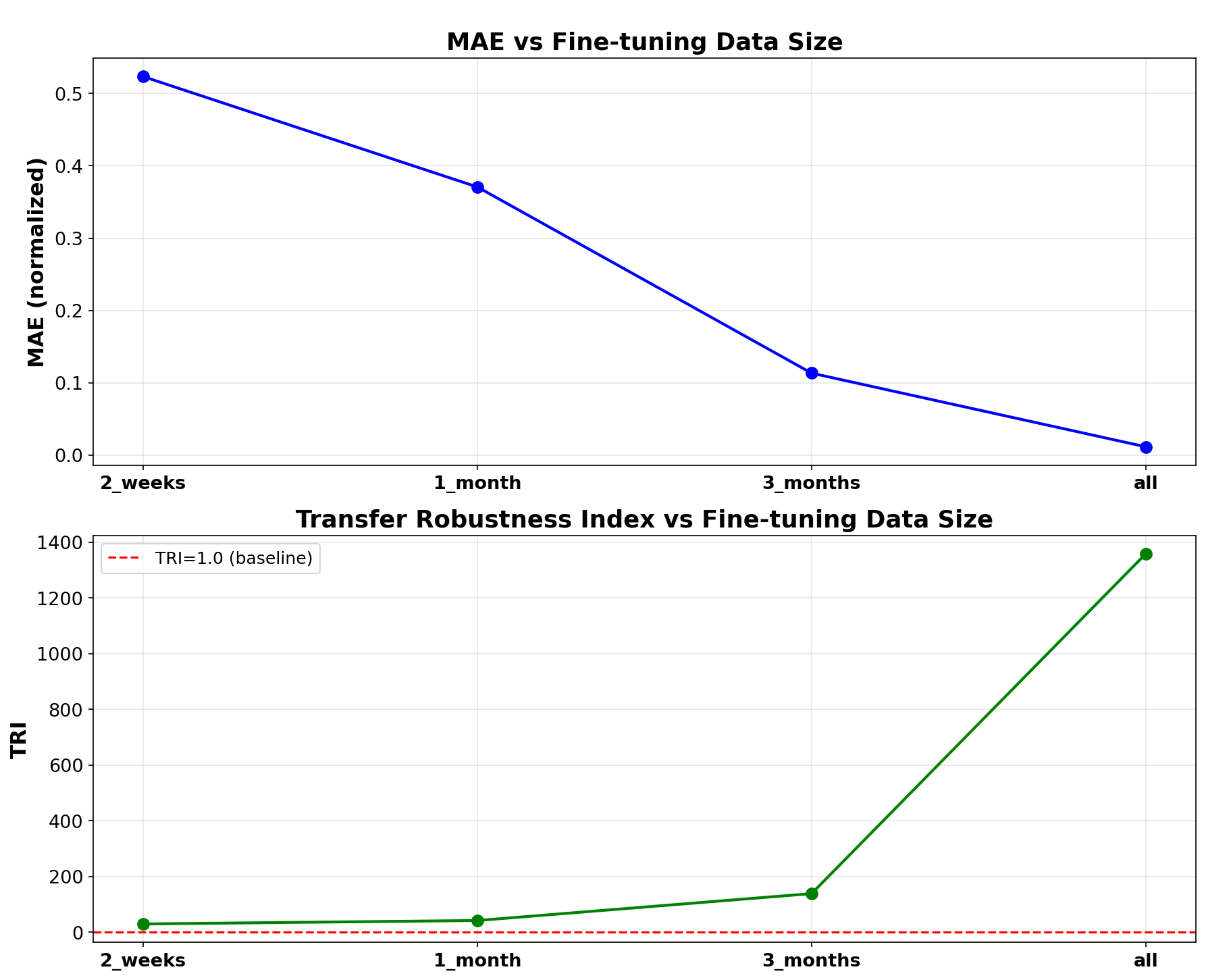}
  \caption{Data-scarcity analysis. Left: test-set MAE versus the size of the NEST fine-tuning window. Right: TRI for the same windows. Performance improves markedly from 3~months onward, with the full pool reaching TRI\,=\,299.}
  \label{fig:scarcity}
\end{figure}

Table~\ref{tab:scarcity} and Figure~\ref{fig:scarcity} illustrate the
impact of target-domain data availability on transfer quality. Note
that the scarcity experiments use a reduced 15-epoch budget and
independent random initialization to simulate rapid deployment
scenarios. While the resulting $\mathit{MAE}$ for the full pool
($0.053$) is higher than the optimized baseline in
Table~\ref{tab:results} ($0.0068$), the monotonic improvement confirms
that data volume is the primary driver of transfer quality regardless
of the training budget.

\begin{table}[t]
\caption{Data scarcity analysis: NEST test performance vs.\
  fine-tuning window. \textbf{Bold} = best.}
\label{tab:scarcity}
\footnotesize
\setlength{\tabcolsep}{4pt}
\begin{tabular}{lrrr}
\toprule
\textbf{Window} & \textbf{Hours} & \textbf{MAE} & \textbf{TRI} \\
\midrule
2 weeks   &   336 & 0.404 &  38.9 \\
1 month   &   720 & 0.399 &  39.4 \\
3 months  & $2{,}160$ & 0.163 &  96.8 \\
\textbf{All data} & \textbf{$4{,}965$} & \textbf{0.053} & \textbf{299.0} \\
\bottomrule
\end{tabular}
\end{table}

Results show that 2~weeks and 1~month perform similarly
(MAE~$\approx$~0.40, TRI~$\approx$~39), while 3~months gives a clear
improvement (MAE\,=\,0.163, TRI\,=\,96.8) and full data performs best
(MAE\,=\,0.053, TRI\,=\,299). Thus, about \emph{3~months} of
target-building data appears necessary for meaningful transfer quality.

\section{Discussion}

\textbf{Why Probe-Only Wins.} Updating only 455 output parameters
beats full fine-tuning, consistent with catastrophic-forgetting
theory: updating all layers on limited target data can overwrite
transferable source representations. The TFT encoder trained on
$13{,}244$~h of source data appears to capture reusable temporal
structures such as occupancy-driven peaks and weekly periodicity,
while only the final mapping to the target distribution requires
adaptation.

\textbf{Scope, Baselines, and TRI.} The study covers a strong domain
gap involving nation, climate, typology, and system differences, but
should still be read as a \emph{single-target case study}. Treating
the four NEST units as independent targets and adding more buildings
is the natural next step. Table~\ref{tab:results} focuses on
TL-relevant comparators---Persistence, LSTM-from-scratch, Direct
Transfer, and four fine-tuning strategies---to isolate the effect of
transfer on the same test set. Wider comparison with N-BEATS, vanilla
Transformer, ARIMA, and Prophet is reserved for extended work. We
report $\mathit{TRI}$ as an open, architecture-agnostic diagnostic for
comparing TL strategies in this case study.

\textbf{Limitations.} Negative $R^2$ on NEST reflects high-frequency
spikes, the November--May vs.\ June--July seasonal shift, and
zero-padding for cross-domain alignment. Future work will consider
domain alignment, conformal prediction, multi-target forecasting, and
larger building portfolios with scale-independent metrics such as
CV-RMSE and MASE.

\section{Conclusion}

We presented an uncertainty-aware TL framework for cross-building
energy forecasting on a newly released sub-meter dataset
(AAU, Denmark $\to$ NEST, Switzerland). The paper makes four
contributions: a \textbf{TFT-based} TL pipeline; the Transfer Robustness Index
as a reusable community metric; a layer-freezing ablation identifying
Probe-Only fine-tuning with 455 trainable parameters as the strongest
strategy ($\mathit{TRI}=3{,}097$); and calibrated MC Dropout
uncertainty ($\mathit{PICP}=93.2\%$) together with a data-scarcity
analysis. The central insight is that TFT encoders trained on
large-scale sub-meter data show promising transfer across typology and
climate differences without full retraining, providing a foundation for
scalable, uncertainty-aware district-level energy management.

{\footnotesize
\bibliographystyle{ACM-Reference-Format}
\bibliography{references}
}

\end{document}